%% file: main.tex
\pdfoutput=1
\documentclass[11pt]{article}

\usepackage[]{ACL2023}
\usepackage{tikz}

\usepackage{times}
\usepackage{latexsym}

\usepackage[T1]{fontenc}
\usepackage[utf8]{inputenc}

\usepackage{microtype}

\usepackage{inconsolata}

\title{Best Preprocessing Techniques for Sentiment Analysis}

\author{
  Saranzaya Magsarjav$^\diamond$ 
  \and
  Melissa Humphries$^\dagger$
  \and
  Jonathan Tuke $^\star$
  \and
  Lewis Mitchell$^\ast$
  \ \\
  \\
  $^\diamond$ $^\dagger$ $^\star$ $^\ast$ The School of Mathematical Sciences,
  \\
  Adelaide University,
  \\
  South Australia 5005, Australia
  \\
  {\small
  $^\diamond$ \texttt{saranzaya.magsarjav@adelaide.edu.au}, $^\dagger$\texttt{melissa.humphries@adelaide.edu.au}}\\
  {\small
  $^\star$\texttt{simon.tuke@adelaide.edu.au}, $^\ast$\texttt{lewis.mitchell@adelaide.edu.au}}
}

\begin{document}

\maketitle

\begin{abstract}
   Sentiment analysis in Twitter datasets is important because it enables monitoring public opinion on products and analysis of political and social movements. One critical step is preprocessing: the automated processing of text for machine learning algorithms. Preprocessing plays a critical role in reducing noise and improving efficiency. However, little research has systematically examined the order in which preprocessing techniques are implemented. We find that, when accounting for order, spelling correction is the least impactful preprocessing technique, whereas tokenisation is the most impactful. Stemming and stop-word removal are interchangeable, and it is better to remove stop words without removing negation. The best order for applying the preprocessing techniques was tokenisation, text cleaning, stemming, and then stopword removal. Our results provide a systematic approach for practitioners to deploy preprocessing to improve model output without the costly preprocessing exploratory phase.
\end{abstract}

\input{introduction.tex}

\input{dataset.tex}

\input{method.tex}

\input{results_V1.tex}
\input{conclusion}

\bibliography{preprocessing_paper}
\bibliographystyle{acl_natbib}

\end{document}

%% file: introduction.tex
\section{Introduction}

Since the seminal work by \citet{pang2008opinion}, Opinion Mining and Sentiment Analysis have grown exponentially. This growth is largely due to the increased availability of social media, which is specifically designed for sharing opinions, views, and experiences. As a result, social media has become a natural fit for Sentiment Analysis. 

Language is context-based, making it extremely difficult to analyse. Sentiment in particular can be very subtle. Things like irony, sarcasm and negation, where a single word can completely change the sentiment polarity, all complicate analysis. These challenges are further compounded when using social media, given the informal nature of online content. Social media posts frequently include URLs, hashtags, multimedia, emoticons, misspellings, and slang, which increase the dimensionality of the problem and make classification more complex. Therefore, before sentiment analysis and classification, the data must undergo carefully considered preprocessing steps to reduce unnecessary noise. 

Preprocessing steps are used to improve the performance of classifiers, whether in Sentiment Analysis or otherwise, by converting text into a more manageable and analysable form. Some basic techniques include stemming (reducing words to their root forms), converting to lowercase, and removing stop words (e.g., pronouns and articles). 

In addition to these relatively well-established techniques, newer preprocessing methods for online content include emoji conversion \cite{dandannavarEmoticonsTheirEffects2020}, slang conversion \cite{singhRoleTextPreprocessing2016}, spelling correction, and many others. However, few papers consider how preprocessing techniques are applied or in what order. No systematic analysis has tested different orders of preprocessing techniques. This paper aims to fill this gap by systematically testing the ordering of preprocessing techniques and identifying which should be implemented. This, in turn, allows us to provide practitioners with recommendations on implementation orders. We show that tokenisation is the most impactful preprocessing technique and spelling correction is the least impactful. The order that yields the best output is tokenisation, cleaning, stemming, and then stopword removal. Although there has been an increase in the use of Neural Network models, such as BERT, in sentiment analysis, this paper will primarily focus on word-based sentiment analysis. The simplicity of the models will help focus on changes to preprocessing techniques rather than on differences between models.

The remainder of this work is organised as follows. Section \ref{sec:related} presents a review of the current literature, and Sections \ref{sec:data} onwards detail the methods, results, and discussion. Finally, the conclusion and future work are in Section \ref{sec:conclusion}.

\begin{figure*}
    \centering
    \includegraphics[width =0.7\textwidth]{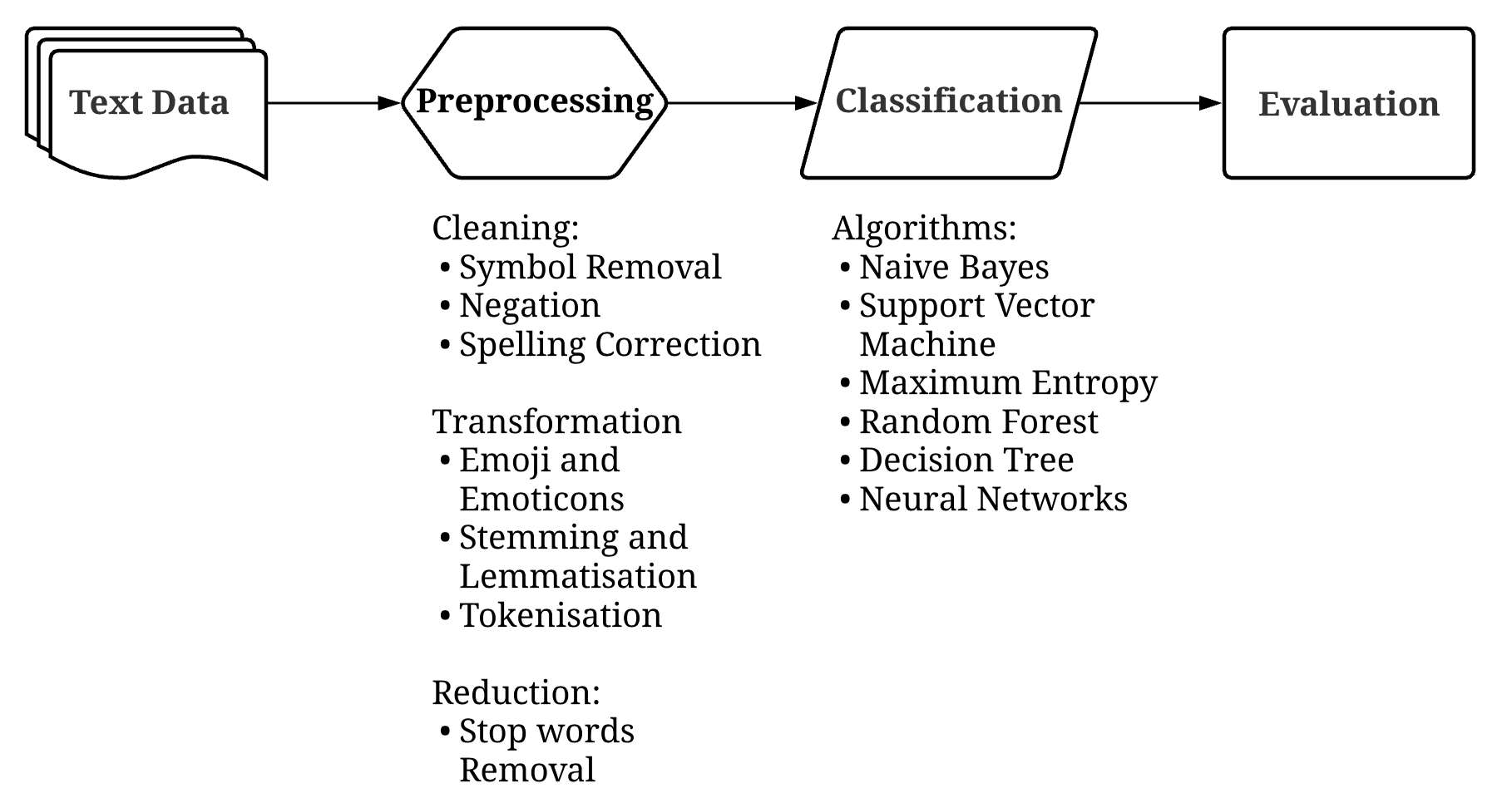}
    \caption{A flowchart process for sentiment analysis classification.}
    \label{fig:flowchart}
\end{figure*}

\section{Related Work} \label{sec:related}

The current literature shows that preprocessing significantly affects sentiment classification. The best methods, however, seem highly dependent on both the algorithm used and the context of the investigation. To our knowledge, no systematic investigation of the ordering of the techniques has been conducted.

\citet{Angiani2016} used Multinomial Naive Bayes on SemEval 2015 \citep{semeval2015} and 2016 \citep{semeval2016} data and showed the individual effects of several preprocessing techniques, including converting all negations to `not', emoji to simple descriptions, spelling correction, slang conversion, stemming, and stopword removal. All preprocessing techniques applied improved the output except for spelling correction and slang conversion. When combined, basic cleaning methods significantly improved classification compared to no cleaning; therefore, they recommended applying these techniques before any other preprocessing.

Expanding on this, \citet{alam2019impact} also used Naive Bayes to assess the effects of different preprocessing steps on the output. They compared this to the effects of preprocessing on Support Vector Machine and Maximum Entropy Modelling. The base comparison used emoticon removal. The authors also applied bi-grams. The most significant improvement was observed with Naive Bayes. However, \citet{alam2019impact} showed this was not true for all algorithms, with maximum entropy showing no improvement in accuracy.

\citet{jianqiang2015pre} showed the effects of URL, stopword, repeated letters, negation, acronym and number removal on sentiment classification performance using two feature models and four classifiers (Logistic regression, Naive Bayes, Support Vector Machines, Random Forests) on five Twitter datasets. When focusing on testing individual preprocessing techniques, there was no consensus on which were best, as accuracy changes varied across datasets and classifiers. The literature presents a wide variety of outcomes; however, only a few examine the simultaneous application of multiple preprocessing techniques, and no papers systematically analyse their ordering. This paper aims to fill this gap in the literature.

\section{Framework and Algorithms} \label{sec:frame}

\subsection{Framework}

A standard supervised sentiment classification process is shown in Figure \ref{fig:flowchart}. After the dataset is collected, it will undergo preprocessing. This step cleans the data and reduces it to the most informative set. This step can also be part of the feature selection process. The selected features and different feature combinations will affect the classifier's performance. From this set, the model is trained, typically using a machine-learning classification algorithm. An overview of the algorithms used is provided in Section \ref{sec:predict}. The classifier then assigns labels, and the predictions are evaluated.

\subsection{Prediction Algorithms}\label{sec:predict}

The main methods used in this paper for text preprocessing are Naive Bayes, Support Vector Machine, Clustering, and Decision Trees. We use these standard techniques for simplicity. Many more methods can be used to classify and evaluate sentiment analysis problems; refer to \citet{yue2019survey} and \citet{Giachanou2016a}. We do not provide an extensive review of prediction algorithms here, but direct the reader to the following resources. \citet{Giachanou2016a} provides a good overview of sentiment analysis and opinion mining and their application to Twitter. It presents many challenges in sentiment analysis and in using Twitter for it, as well as features, applications, and open problems in the field. \citet{yue2019survey} is a more in-depth review of different types of sentiment analysis and their backgrounds. It presents the finer details of sentiment analysis and the different types of sentiment analysis and opinion mining.

\subsection{Preprocessing Techniques}

Data preprocessing has four main steps: cleaning, integration, transformation, and reduction. When applied specifically to text preprocessing, the major components are cleaning, transformation and reduction. These components help reduce dataset volume and noise by normalising, aggregating, or integrating data in various ways. The specific preprocessing techniques that we focused on were:

\begin{itemize}[nosep]
    \item Spelling correction: The process of correcting misspelled words.
    \item Stemming: Reducing words to their root form.
    \item Tokenisation: Segmenting text or strings of characters into paragraphs, sentences, words or characters for easier analysis.
    \item  Stop word removal: Removing common words that do not add to sentiment analysis, such as pronouns and articles.
\end{itemize}

The final preprocessing technique considered was cleaning. The cleaning processes that are considered in this paper are:

\begin{itemize}[nosep]
    \item emoji conversion: converting the emoji/ emoticons to worded descriptions.
    \item converting to lowercase,
    \item  de-contraction: using a list of known contractions and converting them to their expanded form, and
    \item  symbol removal: removal of special characters and URLs
    \item punctuation removal
\end{itemize}

To compare how each preprocessing technique affects outcomes, the option of not implementing any cleaning was also considered. The order in which these were applied was kept the same to reduce computational cost, since the preprocessing techniques are the main focus.

%% file: dataset.tex
\section{Data Set} \label{sec:data}

\begin{table}
\begin{center}
\caption{Final number of positive and negative tweets in each dataset. }
\label{tab:data_num}
\rowcolors{1}{}{lightgray}
\begin{tabular}{|c|c|c|c|} \hline
        & Positive & Negative & Total \\ \hline
Airline & 530    &    1125   &  1655 \\
Debate  & 257     & 1396     & 1653  \\
SMILE   & 1148     & 107     & 1255  \\\hline
\end{tabular}
\end{center}
\end{table}

Three datasets were used: US Airline \cite{airline2016}, GOP Debate \cite{debate2016}, and the SMILE project \cite{Wang2016}. All datasets consist of Twitter posts, focusing the analysis on short-form text. To standardise the datasets, neutral tweets were removed, happy emotions were converted to a single positive value, and negative emotions were converted to a single negative value. This preprocessing ensured consistency across datasets before sampling and analysis. 

% The SMILE project data set was collected using 13 different Twitter handles associated with the British Museum between May 2013 and June 2015. It contains 3,085 tweets with five emotions: anger, disgust, happiness, surprise and sadness. To make the data consistent across all the datasets, negative emotions (sadness, anger, and disgust) were converted to negative, and happiness to positive. This reduces the dataset to 1255 tweets: 1148 positive and 107 negative. 

% The GOP Debate dataset was collected during the GOP debate in Ohio in early August 2015. The dataset contains 13871 tweets with positive, negative, and neutral sentiment with sentiment confidence. The data set was reduced by removing neutral sentiment and sentiment confidence that was not one. The final number was 4725 tweets: 735 positive and 3990 negative tweets. 

% The US Airline dataset was collected using handles of major US airlines in February 2015. The data set has 14640 tweets with labelled positive, negative, and neutral datasets. Neutral tweets and tweets with confidence values other than one were removed. This results in 4729 tweets: 1515 positive and 3214 negative tweets.  

After this standardisation process, stratified sampling was applied. Theoretically, there are approximately 1.5 million possible combinations to run; therefore, stratified sampling was used to improve time efficiency. To balance accuracy and efficiency, different sample sizes were tested across datasets for randomly selected combinations of preprocessing techniques. The data were sampled at different proportions multiple times to obtain a confidence interval for the F1 score. This process showed the appropriate sample size was about \(35\%\) of the original dataset. Larger sample sizes did not show a significant improvement in the classification accuracy. The final number of tweets for each dataset is shown in Table \ref{tab:data_num}.

%% file: method.tex
\section{Method}

\begin{table*}
    \centering
    \caption{The orders in which the preprocessing techniques were implemented. The orders are referenced by the number and shorthand in the first column. In total, 15 different orders were considered.}
    \label{tab:orderDF}
    \rowcolors{1}{}{lightgray}
    % \resizebox{0.9\textwidth}{!}{%

    \begin{tabular}{cccccc}
    0:cl-to-sp-st-se  & clean     & tokeniser & spell & stopword  & stem     \\
    1:cl-to-sp-se-st  & clean     & tokeniser & spell & stem      & stopword \\
    2:cl-to-st-sp-se  & clean     & tokeniser & stopword  & spell & stem     \\
    3:to-cl-sp-st-se  & tokeniser & clean     & spell & stopword  & stem     \\
    4:to-cl-sp-se-st  & tokeniser & clean     & spell & stem      & stopword \\
    5:to-cl-st-sp-se  & tokeniser & clean     & stopword  & spell & stem     \\
    6:to-sp-cl-st-se  & tokeniser & spell & clean     & stopword  & stem     \\
    7:to-sp-cl-se-st  & tokeniser & spell & clean     & stem      & stopword \\
    8:to-sp-st-cl-se  & tokeniser & spell & stopword  & clean     & stem     \\
    9:to-sp-st-se-cl  & tokeniser & spell & stopword  & stem      & clean    \\
    10:to-sp-se-cl-st & tokeniser & spell & stem      & clean     & stopword \\
    11:to-sp-se-st-cl & tokeniser & spell & stem      & stopword  & clean    \\
    12:to-st-cl-sp-se & tokeniser & stopword  & clean & spell & stem     \\
    13:to-st-sp-cl-se & tokeniser & stopword  & spell & clean     & stem     \\
    14:to-st-sp-se-cl & tokeniser & stopword  & spell & stem      & clean   
    \end{tabular}
    % }
\end{table*}

Different packages were also used to see which preprocessing package performs best. The packages used for each preprocessing technique are listed below:

\begin{itemize}[nosep]
    \item Spelling correction: spellchecker \cite{pyspell}, textblob \cite{textblob}, autocorrect \cite{autocorrect},
    \item Stemming: SnowballStemmer \cite{nltk}, WordNetLemmatizer \citet{textblob}, spaCy \cite{spacy}, textblob \cite{textblob},
    \item Tokenisation: TweetTokenizer \cite{nltk}, spaCy \cite{spacy}, transformers AutoTokenizer \cite{tokenizers}, whitespace, and
    \item  Stop word removal: nltk stopword \cite{nltk} without removing no and not from the list. 
    \item Cleaning: emoji conversion \citep{emoji}, converting lowercase, de-contraction, symbol removal, punctuation removal \citet{nltk}
\end{itemize}

The possible order of implementation is \(5!\) for the different preprocessing techniques. Trying to run all possible combinations and orders is computationally intensive. Therefore, the computation was improved by accounting for order limitations and illogical combinations, thereby reducing the number of combinations to consider. Stop word removal, spelling correction, and stemming take in tokenised text; thus, they must be implemented after tokenisation. Text transformation and reduction can be applied before or after tokenisation. As string identification is used for all cleaning processes, the text does not need to be tokenised. Another order implementation considered was to run spelling correction before stemming, since the stemmers do not work on incorrect words. For spelling correction, stemming algorithms, and stop word removal to work, the text has to be tokenised. These factors reduced the number of possible orders from \(5!\) to 15, and the possible orders are in Table \ref{tab:orderDF}.

There are \(2^5\) different combinations for the cleaning process and four different spelling correction algorithms, including no implementation. Similarly, for stemming, five methods were applied, with no-stemming also considered. There were four tokenisation methods, including whitespace, and the stopwords could be removed or not removed. All of these add up to \(2^5\times 4 \times 5 \times 4 \times 2 = 5120\) possible combinations of the different techniques. After cleaning, the final text is run through four different models for sentiment analysis: Naive Bayes (NB), K-means (KM), Decision Trees (DT), and Support Vector Machines (SVM). To mitigate overfitting, five-fold cross-validation was used, and the average F1 scores were recorded, where the F1 score accuracy is defined as:
\[
    F1 = \frac{2 \times TP} {2 \times TP + FN + FP},
\]
where TP is the count of true positives, FN is the count of false negatives, and FP is the count of false positives. Analysis of variance (ANOVA) was used to determine the impact of each process, as it provides information on which techniques had the greatest effect on the outputs.

%% file: results_V1.tex
\section{Results}

\begin{sidewaystable*}
% \begin{table*}
    \centering
    \caption{Average F1-scores for each order for different datasets and for each model. The top 3 highest F1 accuracies are bolded for each model and dataset.}
    \label{tab:order_all}
    \rowcolors{1}{}{lightgray}
    % \resizebox{\textwidth}{!}{
        \begin{tabular}{|c|cccc|cccc|cccc|} \hline
            order(Table \ref{tab:orderDF})
            & \multicolumn{4}{|c|}{airline}                                           & \multicolumn{4}{c|}{debate}                                            & \multicolumn{4}{c|}{SMILE}                                             \\ \hline
           & DT              & KM              & NB              & SVM             & DT              & KM              & NB              & SVM             & DT              & KM              & NB              & SVM             \\ \hline
        0:cl-to-sp-st-se   & \textbf{0.8401} & 0.4566          & 0.8807          & 0.9050          & 0.7707          & \textbf{0.5329} & 0.7744          & 0.8108          & 0.8990          & 0.5587          & \textbf{0.8960} & 0.9126          \\
        1:cl-to-sp-se-st  & \textbf{0.8403} & 0.4576          & 0.8808          & 0.9052          & 0.7707          & 0.5327          & 0.7745          & 0.8110          & 0.8988          & 0.5598          & \textbf{0.8962} & 0.9126          \\
        2:cl-to-st-sp-se  & \textbf{0.8400} & \textbf{0.4592} & 0.8806          & \textbf{0.9052} & 0.7707          & 0.5322          & 0.7739          & 0.8108          & 0.8988          & 0.5596          & \textbf{0.8957} & 0.9125          \\
        3:to-cl-sp-st-se  & 0.8384          & \textbf{0.4595} & \textbf{0.8823} & 0.9051          & \textbf{0.7738} & \textbf{0.5335} & \textbf{0.7776} & \textbf{0.8145} & \textbf{0.8997} & 0.5639          & 0.8941          & \textbf{0.9130} \\
        4:to-cl-sp-se-st  & 0.8386          & 0.4585          & \textbf{0.8824} & \textbf{0.9052} & \textbf{0.7736} & \textbf{0.5338} & \textbf{0.7776} & \textbf{0.8147} & 0.8994          & 0.5655          & 0.8943          & \textbf{0.9130} \\
        5:to-cl-st-sp-se  & 0.8383          & \textbf{0.4598} & \textbf{0.8822} & \textbf{0.9054} & 0.7736          & 0.5312          & \textbf{0.7771} & \textbf{0.8143} & \textbf{0.8995} & 0.5647          & 0.8937          & \textbf{0.9129} \\
        6:to-sp-cl-st-se  & 0.8376          & 0.4555          & 0.8810          & 0.9036          & 0.7720          & 0.5304          & 0.7762          & 0.8121          & 0.8978          & 0.5642          & 0.8934          & 0.9120          \\
        7:to-sp-cl-se-st  & 0.8376          & 0.4542          & 0.8810          & 0.9037          & 0.7720          & 0.5300          & 0.7762          & 0.8122          & 0.8975          & 0.5650          & 0.8936          & 0.9120          \\
        8:to-sp-st-cl-se & 0.8354          & 0.4524          & 0.8792          & 0.9026          & 0.7733          & 0.5207          & 0.7754          & 0.8116          & 0.8985          & 0.5689          & 0.8925          & 0.9119          \\
        9:to-sp-st-se-cl  & 0.8345          & 0.4493          & 0.8784          & 0.9017          & 0.7724          & 0.5219          & 0.7752          & 0.8107          & 0.8974          & \textbf{0.5695} & 0.8916          & 0.9104          \\
        10:to-sp-se-cl-st & 0.8364          & 0.4534          & 0.8801          & 0.9026          & 0.7708          & 0.5296          & 0.7760          & 0.8113          & 0.8965          & 0.5649          & 0.8928          & 0.9104          \\
        11:to-sp-se-st-cl & 0.8347          & 0.4487          & 0.8784          & 0.9019          & 0.7717          & 0.5208          & 0.7754          & 0.8108          & 0.8969          & 0.5694          & 0.8919          & 0.9104          \\
        12:to-st-cl-sp-se & 0.8355          & 0.4561          & 0.8802          & 0.9040          & \textbf{0.7743} & 0.5212          & 0.7763          & 0.8135          & \textbf{0.9000} & \textbf{0.5697} & 0.8926          & 0.9129          \\
        13:to-st-sp-cl-se & 0.8346          & 0.4541          & 0.8792          & 0.9031          & 0.7731          & 0.5205          & 0.7749          & 0.8115          & 0.8980          & 0.5693          & 0.8919          & 0.9120          \\
        14:to-st-sp-se-cl & 0.8338          & 0.4517          & 0.8784          & 0.9021          & 0.7723          & 0.5205          & 0.7747          & 0.8105          & 0.8970          & \textbf{0.5695} & 0.8911          & 0.9105          \\ \hline
        \end{tabular}
        % }
% \end{table*}
\end{sidewaystable*}

Table \ref{tab:order_all} shows the average F1 scores of each order for each dataset and classifier. The Support Vector Machine consistently performed best, achieving an F1 score above \(90\%\). Decision Tree and Naive Bayes performed similarly. Naive Bayes performed better on the Airline dataset, and Decision Trees performed better on the SMILE dataset and yielded very similar results on the Debate dataset. K-means came last as it did not achieve accuracies greater than \(60\%\).

The best ordering was determined by first considering each classifier individually, then collating similarities among the top-performing orders. Firstly, SVM performed best with the following order implementations: 3:to-cl-sp-st-se, 4:to-cl-sp-se-st, and 5:to-cl-st-sp-se. The common factors across these orders are: tokenisation, cleaning, then spelling correction, and no order preference for stemming or stop-word removal. For Naive Bayes, the order that gave the highest accuracy in two of the datasets was the same as SVM: 3:to-cl-sp-st-se, 4:to-cl-sp-se-st, and 5:to-cl-st-sp-se. The other order implementations are 0:cl-to-sp-st-se, 1:cl-to-sp-se-st, and 2:cl-to-st-sp-se. For KM and DT, the cleaning process was much more dependent on the dataset used. KM and DT are both highly sensitive to the starting initial conditions. K-means are highly sensitive to the initial centroid. Because we are varying the starting point for each data set, the initial centroid will differ across cleaning processes. Similarly, Decision Trees are not robust; thus, slight changes in the dataset can alter the branches and, consequently, the final prediction.

The order that resulted in the higher F1 accuracy across the different models and data sets was: tokenising, cleaning, spelling correction, stop word removal, and then stemming. However, when accounting for all variations in the top-order implementation, tokenisation and cleaning processes could be interchangeable; cleaning came before spelling correction, and stemming and stop-word removal were the last steps.

Noting that the cleaning process is more consistent when using SVM, the analysis will focus on SVM across the three datasets.

\begin{figure*}
    \centering
    \includegraphics[width = \textwidth]{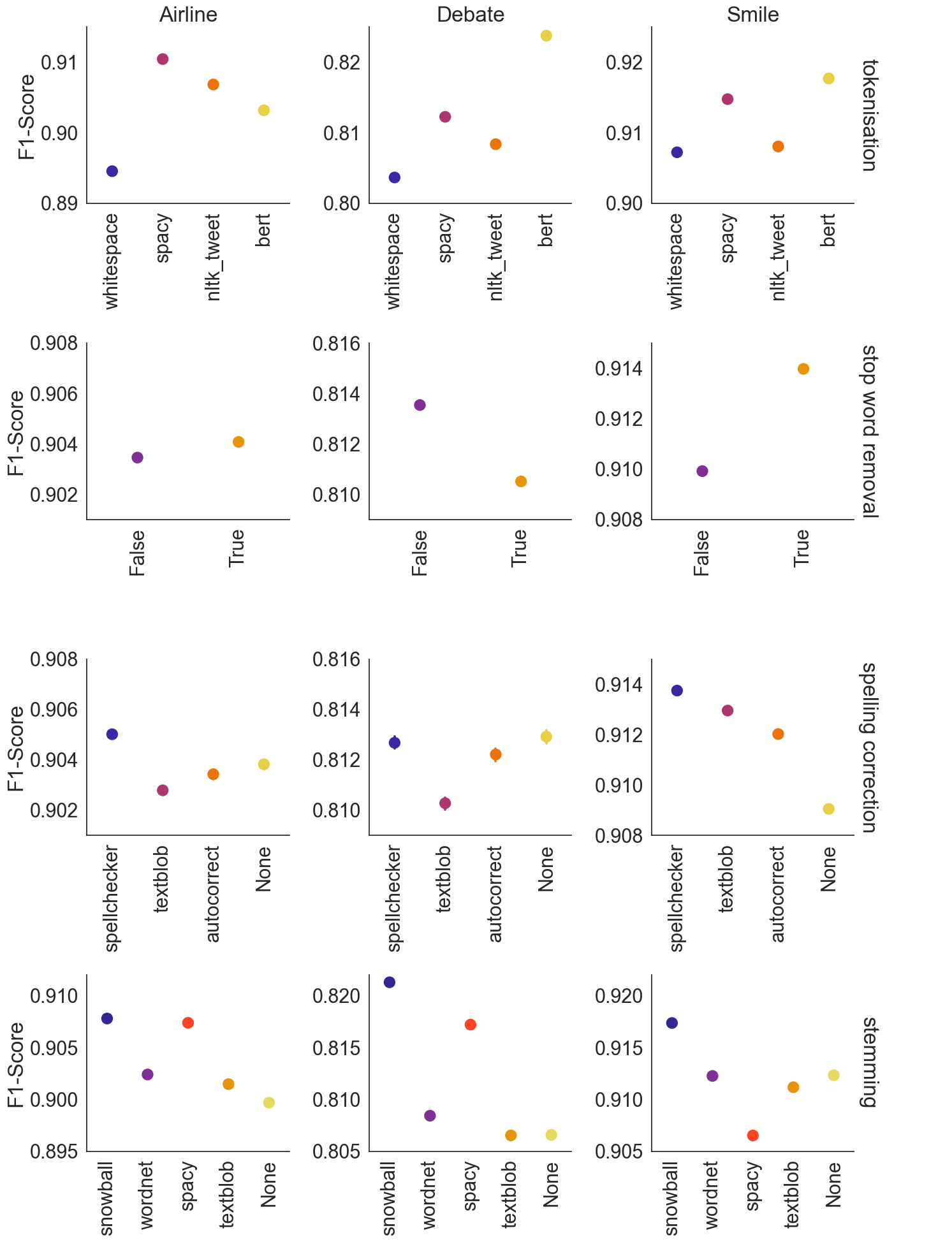}
    \caption{The average F1-scores with standard error bars for different preprocessing techniques. The results are from SVM on the three datasets. The rows are the different techniques, and the columns are the different datasets.}
    \label{fig:prep_SVM}
\end{figure*}

% \textcolor{red}{For \cref{fig:prep_SVM}, I have tried to condense the figure to one row, in \cref{fig:SVM_prep_max} and in \cref{fig:SVM_prep_all}. 
% For \cref{fig:SVM_prep_max}, I just chose one of the packages, that was performing the best and compared that to not applying the technique. For \cref{fig:SVM_prep_max}, I put all the packages into one big `yes' category. Not sure if this is necessary or if it adds too much ambiguity. Need some guidance?}

\begin{table}
    \centering
    \caption{ANOVA outputs for the different datasets using SVM. A higher F-statistic indicates greater variation in the results; therefore, it has a greater impact on the output. It can be observed that tokenisation is the most impactful, and spelling correction is the least impactful.}
    \label{tab:svm_anova}
    \resizebox{0.45\textwidth}{!}{
        \begin{tabular}{llllll} \hline
SVM              &           & SS       & DF   & F     & PR(\textgreater{}F) \\ \hline
\multirow{6}{*}{Debate} & \cellcolor{lightgray} tokeniser & \cellcolor{lightgray} 4.246   &\cellcolor{lightgray}  3     & \cellcolor{lightgray}13946 &\cellcolor{lightgray}\(<2\times 10^{-16}\)      \\ 
                     & stem      & 2.849   & 4     & 7020  & \(<2\times 10^{-16}\)                  \\
                     &\cellcolor{lightgray} clean     & \cellcolor{lightgray}8.346   &\cellcolor{lightgray} 31    & \cellcolor{lightgray}2653  & \cellcolor{lightgray}\(<2\times 10^{-16}\)                    \\
                     & stopword  & 0.177   & 1     & 1739  & \(<2\times 10^{-16}\)                 \\
                     &\cellcolor{lightgray} spell     & \cellcolor{lightgray} 0.083   & \cellcolor{lightgray} 3     & \cellcolor{lightgray} 272   & \cellcolor{lightgray} \(<2\times 10^{-16}\)             \\
                     & Residual  & 7.789   & 76757 &       &                    \\ \hline
\multirow{6}{*}{Airline} & \cellcolor{lightgray}tokeniser & \cellcolor{lightgray}2.676    & \cellcolor{lightgray}3     & \cellcolor{lightgray}13978 & \cellcolor{lightgray}\(<2\times 10^{-16}\)                     \\
                     & stem      & 0.816    & 4     & 3196  & \(<2\times 10^{-16}\)                 \\
                     & \cellcolor{lightgray} clean     & \cellcolor{lightgray} 1.101    & \cellcolor{lightgray} 31    &\cellcolor{lightgray}  557   &\cellcolor{lightgray}  \(<2\times 10^{-16}\)                     \\
                     & spell     & 0.051    & 3     & 264   &\(<2\times 10^{-16}\)           \\ 
                     & \cellcolor{lightgray} stopword  & \cellcolor{lightgray}0.007    &\cellcolor{lightgray} 1     & \cellcolor{lightgray}116   & \cellcolor{lightgray}\(<2\times 10^{-16}\)             \\
                     & Residual  & 4.898    & 76757 &       &                    \\ \hline
\multirow{6}{*}{SMILE} & \cellcolor{lightgray}tokeniser &\cellcolor{lightgray} 1.504   &\cellcolor{lightgray} 3     & \cellcolor{lightgray}\cellcolor{lightgray}10354 & \(<2\times 10^{-16}\)                    \\
                     & stopword  & 0.316   & 1     & 6533  &\(<2\times 10^{-16}\)                    \\
                     & \cellcolor{lightgray} stem      &\cellcolor{lightgray} 0.916   &\cellcolor{lightgray} 4     &\cellcolor{lightgray} 4730  &\cellcolor{lightgray} \(<2\times 10^{-16}\)                    \\
                     & spell     & 0.243   & 3     & 1673  &\(<2\times 10^{-16}\)                  \\
                     &\cellcolor{lightgray} clean     & \cellcolor{lightgray}1.317   & \cellcolor{lightgray}31    & \cellcolor{lightgray}878   & \cellcolor{lightgray}\(<2\times 10^{-16}\)                   \\
                     & Residual  & 3.715   & 76757 &       &                    \\ \hline
        \end{tabular} 
        }
\end{table}

\begin{figure*}
    \centering
    \includegraphics[width=0.85\textwidth]{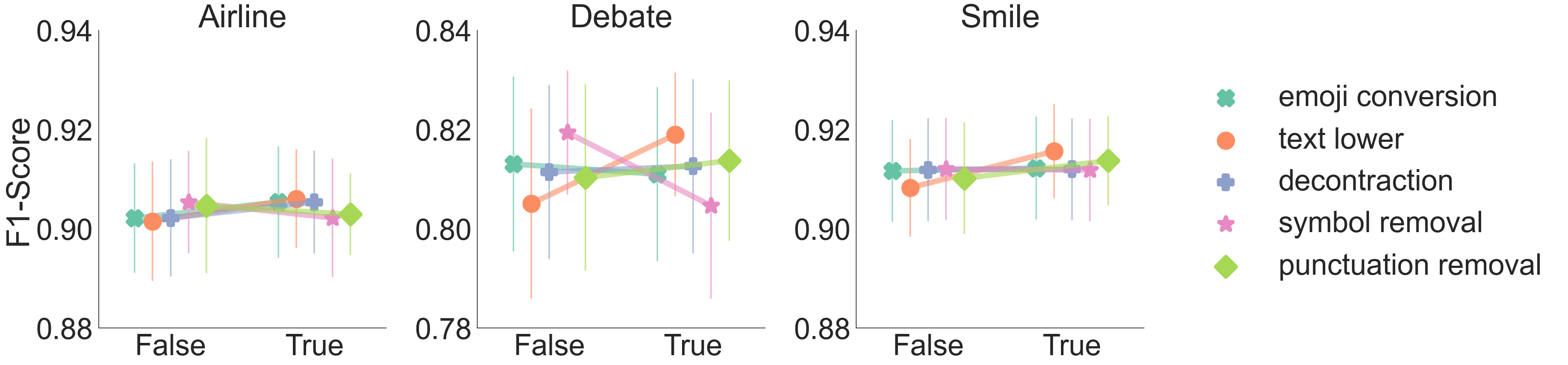}
    \caption{The average F1-score of the cleaning process on different datasets using SVM. The error bar shows the spread of the dataset. If the line between two points is positive, the cleaning process should be applied; if it is negative, it should not be applied.}
    \label{fig:prep_SVM_clean}
\end{figure*}

Table \ref{tab:svm_anova}  shows the ANOVA output and the F-statistic for the different preprocessing techniques for each dataset using SVM. The larger the F-statistic, the more it explains the variation in the data, and the more impactful the preprocessing is on the final output. From the ANOVA outputs, it can be seen that spelling correction has the lowest F-statistic across all datasets and models, with Values ranging from 18 to 1000. This implies that spelling correction is the least impactful preprocessing technique. This was also true when accounting for the order of application. The top performing spelling correction methods, from Figure \ref{fig:prep_SVM}, were spellchecker or no applications of spelling correction, then autocorrect and lastly textblob. 

The most impactful preprocessing technique is tokenisation, which had F-statistics at least an order of magnitude higher than those for spelling correction. The most frequently top-performing methods in Figure \ref{fig:prep_SVM} were BERT and spaCy tokenisers, followed by nltk\_tweet, then whitespace. 

The next couple of important features are stemming and stop word removal. These two preprocessing techniques were interchangeable in most cases. The best-performing stemmer is the snowball stemmer or spaCy stemmer, seen in Figure \ref{fig:prep_SVM}. The Snowball stemmer follows simple rules as a decision process, while the spaCy stemmer takes context into account to determine a word's root. From Figure \ref{fig:prep_SVM}, it is better to remove stop words.

Figure \ref{fig:prep_SVM_clean} shows the results of each cleaning process. The only cleaning process that consistently showed a major improvement was text lowering. The other consistent cleaning process was not removing symbols; however, the results showed a smaller difference. The final consistent cleaning process was the de-contraction of text.

\section{Discussion}

Spelling correction was the least useful technique. This could be because these methods do not account for context. As a result, correcting spelling may introduce as much noise as it removes. The recommendation is not to apply spelling correction due to its low overall impact and poor performance. 

The preprocessing technique that made the most difference in the output was tokenisation. This is intuitive, as it creates a basis for how each following technique will be applied. Spelling correction, stemming, and stop word removal methods will not work correctly if punctuation or emojis remain in the tokenised text. Both best-performing tokenisers, BERT and spaCy, carefully consider how to tokenise text. This preserves important context necessary for sentiment analysis. Therefore, it is recommended to use tokenisation methods that can handle the differences between punctuation and emojis.

Stop words should be removed, as they do not carry much meaning. However, in sentiment analysis, keep negating words such as not and no. Negating words flip the sentiment if removed. 

There was no particular order for whether stop-word removal or stemming was used first. This could be because they do not depend on each other. Stemming a stop word will return a stop word; therefore, no real impact is noted. The recommendation here is to perform stop-word removal first, followed by stemming, as stemming a stop word is redundant. Removing stop words first will save time during stemming. 

During the cleaning process, the only major effective step was text lowering. Lowering the text will reduce the number of variations of the same word, therefore, resulting in more consistent data for modelling. Another, however, less effective cleaning process, was not removing symbols. More often than not, emojis convey the sentiment of the text; therefore, removing symbols may be detrimental to sentiment analysis, as emojis can be made from symbols. The final effective cleaning process was de-contraction.  Similar to symbol removal, the change in accuracy was relatively small; however, applying de-contraction helps reduce data variation. This will only work if part of the stop word removal process is not to remove negating words. The other two cleaning processes, emoji conversion and punctuation removal, were much more dataset-dependent, and there was no clear consensus on whether to use them.

We can now propose a recommendation for applying the different preprocessing techniques in the correct order. The first preprocessing technique is to use a context-aware tokeniser, such as BERT or spaCy. When tokenisation is done well, the following preprocessing steps become more consistent and perform well. The next technique to apply is cleaning the tokenised text. The main cleaning steps to implement are text lowering and decontraction. Both of these help reduce data variation. However, whether to use other cleaning processes depends heavily on the dataset. After cleaning the tokenised text, it should be run through a stemmer. The suggested stemmers are the Snowball and spaCy stemmers. Then, stop-word removal should be applied; however, words such as not and no should be retained in the dataset, as they alter the text's sentiment. 

%% file: conclusion.tex
\section{Conclusion} \label{sec:conclusion}

Sentiment analysis has been extensively applied to social media data, and preprocessing techniques are used to improve F1 accuracy. However, there is no consensus on the order or which techniques should be implemented. In this paper, the physical and algorithmic constraints of preprocessing techniques were considered, thus limiting the possible logical order. The reduced number of preprocessing orders was tested to determine the optimal implementation order using three classifiers across three datasets. 

Average F1-score accuracies for each order showed that the best-performing order was tokenisation, cleaning, spelling correction, then stop word removal, and finally stemming. ANOVA analysis also showed that the most impactful preprocessing technique was tokenisation, with BERT and spaCy being the better choices. Spelling correction was the least effective, so it was not implemented. For the different cleaning processes, the largest change was observed with text lowering, and to a lesser extent with word de-contraction. However, the rest of the cleaning process was highly dependent on the classifier and dataset being used. 

Finally, since the start of this analysis, models regarding language and text have improved significantly due to developments in Large Language Models. These models have changed the scope of how natural language processing is approached. Therefore, for future work, it would be interesting to examine how preprocessing affects outcomes in these models and whether preprocessing of text is necessary for language-related tasks.

% \newpage

% \begin{figure*}
%     \centering
%     \includegraphics[width = \textwidth]{SVM_prep_max_tech.png}
%     \caption{The average F1-scores for each preprocessing technique. `None' is indicative of not applying the technique, `yes' is the application of one of the top performing functions. In tokenisation, `none' was considered whitespace tokenisation, and spacy was used to compare how using a non base tokensation method compares. For stemming, snowball stemmer was used to compare to no stemming. Spellchecker was used for spelling correction to compare to no application.}
%     \label{fig:SVM_prep_max}
% \end{figure*}

% \begin{figure*}
%     \centering
%     \includegraphics[width = \textwidth]{SVM_prep_all_tech.png}
%     \caption{
%         The average F1-scores for each preprocessing technique. `None' is indicative of not applying the technique, `yes' is the application of all other techniques tested, for example, in tokenisation, `none' was considered whitespace tokenisation, and all other three tokenisation methods were condensed into `yes'.
%     }
%     \label{fig:SVM_prep_all}
% \end{figure*}